\documentclass{isprs} %
\usepackage{setspace}
\usepackage{geometry} %
\usepackage{epstopdf}
\usepackage[labelsep=period]{caption}  %
\usepackage[british]{babel} 
\usepackage[hang]{footmisc}
\usepackage{algorithm}
\usepackage{algpseudocode}
\usepackage{graphicx} %
\usepackage{booktabs} %
\usepackage{caption} %
\usepackage{booktabs} %
\usepackage{array}    %
\usepackage{float}
\usepackage{subcaption} %
\usepackage{xcolor}
\usepackage[%
  breaklinks=true,
  letterpaper=true,
  colorlinks,
  bookmarks=false,
]{hyperref}
\hypersetup{linkcolor={black}, urlcolor={blue}, citecolor={black}, colorlinks=true}

\usepackage{amsmath}
\usepackage{amssymb}

\usepackage[%
  xindy,
  acronym,
]{glossaries}
\glsdisablehyper
\usepackage{subcaption}
\usepackage{natbib}

\usepackage{siunitx}
\usepackage{hyperref}
\newacronym{gl:LiDAR}{LiDAR}{Light Detection and Ranging}
\newacronym{gl:ViT}{ViT}{Vision Transformer}
\newacronym{gl:DDPM}{DDPM}{Denoising Diffusion Probabilistic Models} 
\newacronym{gl:CDPM}{CDPM}{Centered denoising Diffusion Probabilistic Models} 
\newacronym{gl:CNN}{CNN}{Convolutional Neural Network}
\newacronym{gl:MAE}{MAE}{Masked Autoencoders}
\newacronym{gl:CD}{CD}{Chamfer Distance}
\newacronym{gl:VAE}{VAE}{Variational Autoencoder}
\newacronym{gl:GAN}{GAN}{Generative Adversarial Network}
\newacronym{gl:PVCNN}{PVCNN}{Point-Voxel CNN}

\newcommand{\pc}{PC² }
\newcommand{\ccd}{CCD-3DR }

\newcommand{\papernamedot}{aim2pc}
\newcommand{\papername}{\textsc{\papernamedot\space}}

\usepackage[capitalize]{cleveref} 

\begin{document}

\title{AIM2PC: Aerial Image to 3D Building Point Cloud Reconstruction}
\date{}

 \author{
  Soulaimene Turki \textsuperscript{1,3}, Daniel Panangian \textsuperscript{1}, Houda Chaabouni-Chouayakh  \textsuperscript{2}, Ksenia Bittner \textsuperscript{1} }

\address{
	\textsuperscript{1 }Remote Sensing Technology Institute, German Aerospace Center (DLR), Wessling, Germany - \\
    (soulaimene.turki, daniel.panangian, ksenia.bittner)@dlr.de\\
	\textsuperscript{2 } Sm@rts Laboratory, Digital Research Center of Sfax, Tunisia – houda.chaabouni@crns.rnrt.tn\\
	\textsuperscript{3 }Higher School of Communication of Tunis, Tunisia – soulaimene.turki@supcom.tn\\
}

\abstract{
Three-dimensional urban reconstruction of buildings from single-view images has attracted significant attention over the past two decades. However, recent methods primarily focus on rooftops from aerial images, often overlooking essential geometrical details. Additionally, there is a notable lack of datasets containing complete 3D point clouds for entire buildings, along with challenges in obtaining reliable camera pose information for aerial images. This paper addresses these challenges by presenting a novel methodology, \textsc{\papername}, which utilizes our generated dataset that includes complete 3D point clouds and determined camera poses. Our approach takes features from a single aerial image as input and concatenates them with essential additional conditions, such as binary masks and Sobel edge maps, to enable more edge-aware reconstruction. By incorporating a point cloud diffusion model based on \gls{gl:CDPM}, we project these concatenated features onto the partially denoised point cloud using our camera poses at each diffusion step.
The proposed method is able to reconstruct the complete 3D building point cloud, including wall information and demonstrates superior performance compared to existing baseline techniques.
To allow further comparisons with our methodology the dataset has been made available at \href{https://github.com/Soulaimene/AIM2PCDataset}{https://github.com/Soulaimene/AIM2PCDataset}.
}

\keywords{Point Cloud, Camera Pose, Building Reconstruction, Aerial Image, Single-View Geometry, Diffusion Models}

\maketitle

\sloppy

\glsresetall
\section{Introduction}\label{sec:Introduction}

3D building point clouds are essential for various applications, including navigation, urban planning, and the development of 3D city maps~\citep{biljecki2015applications}. 
To support these uses, accurate and up-to-date 3D building point clouds are crucial. 
Traditionally, techniques like multi-view stereo imagery \citep{duan2016towards} and \gls{gl:LiDAR} \citep{985731} have been employed for their precision. \Gls{gl:LiDAR} which employs aerial platforms like airplanes or drones equipped with laser scanners, offers detailed 3D point coordinates of terrain. However, this technique is not only expensive but can also yield incomplete 3D point cloud of buildings. Typically, a single scan may primarily capture the roof and some wall structure, leaving other critical elements underrepresented or missing.

Multi-view reconstruction, which captures multiple images from different angles, also faces challenges. Satellite-based image capture, commonly used in this approach, suffers from delays due to fixed satellite orbits. Consequently, obtaining new images of a location can take days or weeks, during which significant changes, such as construction or natural disasters, may occur, leading to data gaps. Moreover, high-resolution satellites, essential for detailed reconstructions, often have longer revisit times, further constraining the availability of up-to-date, high-quality imagery.
In response to these limitations, monocular 3D building reconstruction has emerged as a cost-effective and scalable alternative. This technique employs deep learning to create 3D point cloud from single images, offering improved efficiency and accessibility. While generative models have shown success in producing 3D point clouds from single images, they often rely on training data from synthetic or symmetric objects, which can limit their effectiveness in real-world applications.

Methods such as \pc \citep{melas2023pc2} and \ccd \citep{di2023ccd}, which serve as our baseline approaches, have demonstrated promising results in certain 3D reconstruction tasks; however, they also face significant challenges. These methods depend heavily on accurate camera poses for reconstruction, which can be difficult to obtain for aerial images. As a result, their applicability in remote sensing tasks is limited. Since these methods are not designed or trained for aerial data, they often produce incomplete point clouds, missing important structural details like sharp and accurate edges.
\begin{figure}[t]
  \centering
  \subcaptionbox{\label{fig:eginput}}
                {{\includegraphics[width=.28\linewidth]{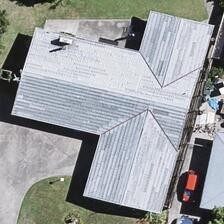}}
                }
  \subcaptionbox{\label{fig:egfig1}}
                {{\includegraphics[width=.28\linewidth]{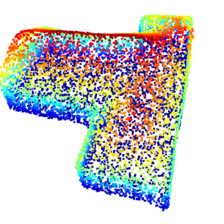}}
                }  
  \subcaptionbox{\label{fig:egfig2}}
                {{\includegraphics[width=.28\linewidth]{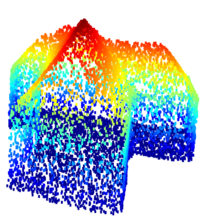}}
                }

  \caption{ 
  Example of the new views (\ref{fig:egfig1}) and (\ref{fig:egfig2}), generated by \papername using the single-view aerial image (\ref{fig:eginput}).
  }
  \label{fig:Intro}
\end{figure}

Additionally, current methods for 3D building reconstruction, such as sat2pc \citep{rezaei2022sat2pc}, focus solely on generating point clouds of rooftops from single-view images, leading to point clouds that lack crucial details and fail to capture the full complexity of the structures.
These methods do not capture the entire building structure from a single view. 
To the best of our knowledge, no existing single-view approaches based on aerial images are capable of reconstructing an entire building, highlighting a critical gap that remains unresolved in the field.
This paper addresses these challenges by investigating the generation of comprehensive building point clouds from single-view aerial images, aiming to enhance the detail and accuracy of 3D point clouds.

In summary, our contributions are as follows:
\begin{enumerate}
  \item We achieve complete building point cloud reconstruction from a single top-view image, encompassing not only the roof point cloud but also the walls and ground as demonstrated in~\cref{fig:Intro}.
  
  \item  We introduce an edge-enhanced method for generating point clouds from single-view aerial images, specifically designed to enhance accuracy and detail in building reconstructions.
  
  \item We generate a dataset containing a complete normalized building point cloud along with associated camera poses.

\end{enumerate}

\section{Related Work}\label{sec:RelatedWork}
\subsubsection*{Single-View 3D Reconstruction.} The challenge of reconstructing 3D shapes from single-view images has been a significant area of interest in computer vision research for over two decades~\citep{fan2017point,chang2015shapenet}. Early approaches primarily focused on extracting visual cues from images, including shading ~\citep{article}, texture \citep{WITKIN198117}, and silhouettes \citep{inproceedings}, to guide the reconstruction process.

As research advanced, a new wave of exploration emerged with the introduction of learning-based methods. Techniques such as those proposed in \citep{Johnston_2017_ICCV,choy20163d} employed encoders to extract 2D features from images, which were subsequently decoded into 3D shapes.

Following these developments, generative models brought a significant transformation in the field. \cite{10.5555/3157096.3157106} applied \glspl{gl:VAE} \citep{kingma2013auto} and \glspl{gl:GAN} \citep{NIPS2014_5ca3e9b1} for 3D shape reconstruction from single images, enabling the generation of diverse and realistic 3D shapes. On the other hand, \cite{achlioptas2018learning} trained an autoencoder to reduce the dimensionality of point clouds and then used a \gls{gl:GAN} to generate 3D shapes from the compressed embeddings. 

Recent developments in 3D single-view reconstruction have led to the introduction of diffusion models ~\citep{ho2020denoising}. For example, \cite{xu2023neurallift} employs a pre-trained depth estimator alongside 2D diffusion priors to recreate the coarse geometry and textures of 3D objects based on a single image. Additionally, \pc \citep{melas2023pc2} and \ccd \citep{di2023ccd}  utilize camera pose to project 2D image features into the point cloud during the reverse diffusion process.

Despite the progress made, most methods tend to  focus on specific 3D benchmarks, such as ShapeNet ~\citep{chang2015shapenet}, which predominantly features synthetic objects characterized by simple and symmetrical shapes. 
\subsubsection*{3D Building Reconstruction.}
Building reconstruction methods have historically relied on the identification of low-level features, such as lines and roof segments, which are subsequently integrated to create the complete roof structure \citep{1641024,s90806101}.
In addition to these geometric features, using shadow information has significantly improved the accuracy of building reconstruction. By analyzing the shadows cast by buildings, researchers can infer height information and better understand the spatial relationships between structures \citep{articlea,323811}.
However, these approaches typically rely on simple building point clouds like squares and rectangles, which reduces their effectiveness for complex designs. Furthermore, the dependence on accurate solar positioning can result in errors under different lighting conditions, limiting their practical use.

Recently, deep learning-based methods have been developed to estimate building height from single 2D images, integrating this with building footprint extraction from nadir-view and near-nadir-view images \citep{li20243d,Mahmud_2020_CVPR}.
The latest findings have introduced sat2pc \citep{rezaei2022sat2pc}, which reconstructs building roofs from single-view images using the GraphX-Conv \citep{nguyen2019graphx} architecture for point cloud deformation, followed by a refinement network. However, this method is limited to rooftops and fails to capture finer details. Another recent approach, BuilDiff \citep{wei2023buildiff}, employs two conditional diffusion models, one for generating coarse representations and the other for creating finer representations from general view images. Despite its innovations, this method struggles to produce accurate 3D models of buildings.

In this paper, we introduce \textsc{\textbf{\papernamedot}}, a novel approach capable of reconstructing an accurate and detailed 3D \textbf{P}oint \textbf{C}loud of an entire building from a single \textbf{A}erial \textbf{Im}age.

\section{Methodology}\label{sec:Methodology}
In the following sections,  we present our methodology, beginning with a brief overview of diffusion models, highlighting the differences between~\gls{gl:DDPM} and~\gls{gl:CDPM}. We then showcase the adopted feature projection method before providing a comprehensive explanation of our edge-enhanced pipeline, \textsc{\papername}, for 3D building reconstruction from single-view aerial images. Additionally, we will detail the process of dataset generation, which is specifically designed to meet our research objectives.

\subsection{Diffusion Models}
In machine learning, diffusion models represent a novel method for generative tasks by reversing the noise addition process. These models are based on the principle of progressively converting a random noise distribution into organized data, like images or 3D point clouds. This transformation occurs in two phases: the forward diffusion process and the reverse denoising process.
The \textbf{forward process} involves the gradual addition of Gaussian noise to clean data \( x_0 \), transforming it into a noisy version \( x_t \) at each time step \( t \), where \( t \in \{1, 2, \dots, T\} \), with \( T \) denoting the total number of diffusion steps. As \( t \) progresses, the data becomes increasingly noisy, ultimately reaching \( x_T \), a fully noisy state resembling a sample from a Gaussian distribution.
This is represented by the following transition:
\begin{equation}
q(x_t | x_{t-1}) = \mathcal{N}(x_t; \sqrt{\alpha_t} x_{t-1}, (1 - \alpha_t) I),
\end{equation}
where $x_t$ is the noisy data at time $t$, $\alpha_t$ is a parameter controlling the noise schedule, and $I$ is the identity matrix.
The full forward process can be condensed as:
\begin{equation}
\label{eq:full_forward}
x_t = \sqrt{\bar{\alpha}_t} x_0 + \sqrt{1 - \bar{\alpha}_t} \epsilon, \quad \epsilon \sim \mathcal{N}(0, I)
\end{equation}
where \( \bar{\alpha}_t = \prod_{s=1}^{t} \alpha_s \), and \( \epsilon \) represents the Gaussian noise added to the data. This process ultimately transforms the clean data \( x_0 \) into a fully noisy version \( x_T \), which resembles a sample from a Gaussian distribution.
The \textbf{reverse process} aims to reconstruct the clean data by removing the noise, step by step, starting from the noisy data \( x_T \). The reverse process is modeled as:
\begin{equation}
p_\theta(x_{t-1} | x_t) = \mathcal{N}(x_{t-1}; \mu_\theta(x_t, t), \Sigma_\theta(t))
\end{equation}
where
\begin{equation}
\mu_\theta(x_t, t) = \frac{1}{\sqrt{\alpha_t}} \left( x_t - \frac{\beta_t}{\sqrt{1 - \bar{\alpha}_t}} \epsilon_\theta(x_t, t) \right)
\end{equation}
is the predicted mean at time step \( t \), with $\epsilon_\theta(x_t, t)$ representing the noise predicted by the neural network, and $\beta_t = 1 - \alpha_t$ denoting the noise removal rate.

This iterative process seeks to recover \( x_0 \) from \( x_T \) by progressively denoising each step. However, when working with point clouds, each point is processed independently. This means that the denoising of one point does not consider the spatial relationships or geometric configurations of other points within the cloud. As a result, there is no geometric prior to ensure that the center of the point cloud remains stable throughout the denoising process. This leads to a problem known as \textbf{center bias} in~\gls{gl:DDPM}~\citep{ho2020denoising}.

In~\gls{gl:CDPM} \citep{di2023ccd}, the goal is to address the center bias that occurs during the reverse process in~\gls{gl:DDPM}. In traditional~\gls{gl:DDPM}, the point cloud's center may shift as the reverse process proceeds, affecting the accuracy of 3D reconstructions. \Gls{gl:CDPM} introduces a novel approach by constraining the denoised point cloud to remain \textbf{zero-mean} at every step of the reverse process, eliminating this center drift.

In practice,~\gls{gl:CDPM} ensures that both the added noise during the forward process and the predicted noise during the reverse process are centered at each step by applying a centralizing operation that adjusts the noise by subtracting its mean, effectively centering it around zero, given by \(\epsilon = \epsilon - \bar{\epsilon}\) and \(\hat{\epsilon} = \hat{\epsilon} - \bar{\hat{\epsilon}}\); similarly, during inference, the point cloud \( x_{t-1} \) is centered as \( x_{t-1} = x_{t-1} - \bar{x}_{t-1} \).

This centering mechanism ensures that the denoised point cloud remains aligned with its original center throughout the reverse process, improving reconstruction quality.
\begin{figure*}[ht]
		\includegraphics[width=1\textwidth]{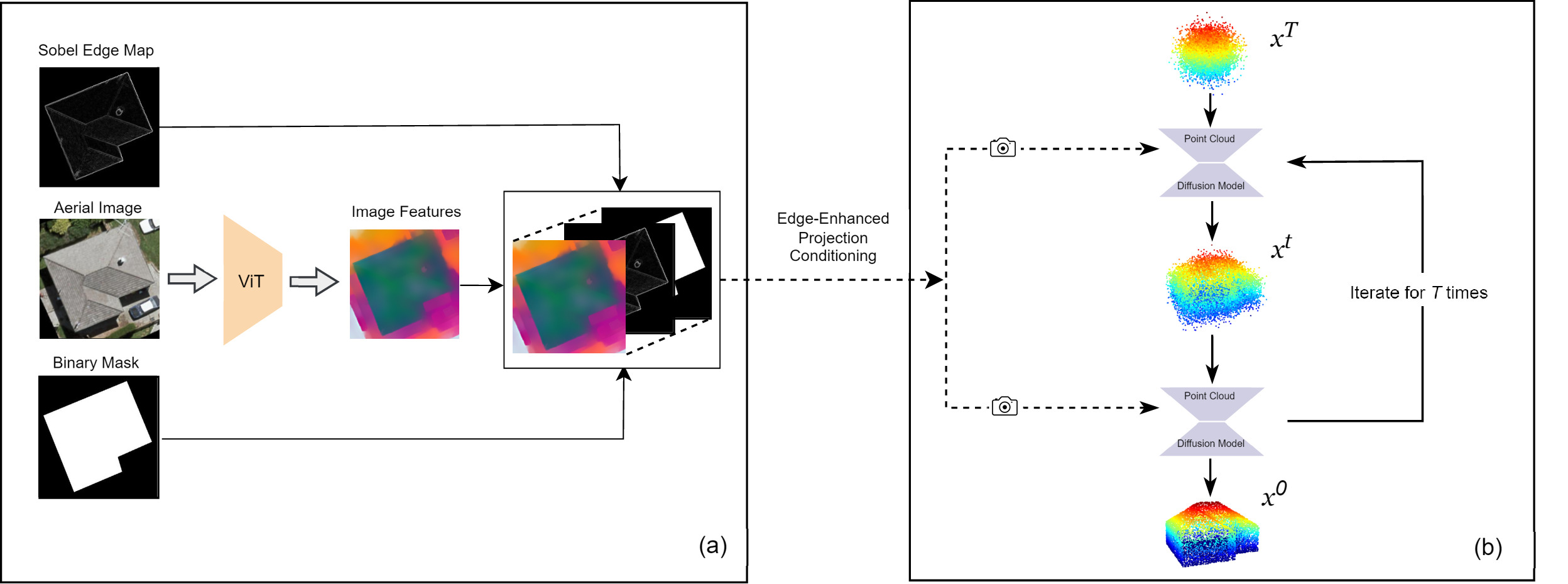}
\caption{Architecture of \papername for reconstructing an entire building from a single aerial image using an edge enhancement approach to effectively capture edges and finer building details. Block (a) illustrates the exact features projected, using the generated camera pose, onto the partially denoised point cloud. These features are a concatenation of those extracted from the ViT of the input image, the binary edge mask, and Sobel edge maps derived from the Sobel operator. Block (b) outlines our reverse process, which begins with Gaussian noise and progressively refines the point cloud to reconstruct the building utilizing a Centered Denoising Probabilistic Model (CDPM), conditioned on the combined features from Block (a).}
\label{fig:architecture}
\end{figure*}
\subsection{Features Projection in Point Cloud Diffusion Models}
In this paper, we consider a point cloud with \( N \) points, represented as a \( 3N \)-dimensional object in space. Our objective is to apply a diffusion model to iteratively denoise this Gaussian point cloud, often visualized as a spherical Gaussian ball, until we achieve a recognizable building point cloud.

The model \( S_\theta \) predicts the offset of each point from its current noisy position \( x_t \). This offset represents how much each point's position deviates from its ideal clean version. 

We iteratively sample \( x_{t-1} \) from the reverse diffusion process until we converge to our target distribution \( q(x_0) \). This process involves minimizing the $\mathcal{L}_2$ loss between the actual noise values and the predicted noise values. Mathematically, this can be expressed as:
\begin{equation}
\mathcal{L}_{2} = \mathbb{E}_{\epsilon} \left[ \| \epsilon - S_\theta(x_t, t) \|_2^2 \right],
\end{equation}
where \( \epsilon \) is the actual noise and \( S_\theta(x_t, t) \) is the model's prediction of the noise at time step \( t \). 

When it comes to conditioning \pc, \cite{melas2023pc2} proposed a method that involves projecting image features onto the partially denoised 3D points at each step of the diffusion process. This process is straightforward: starting with the partially denoised 3D points \( x_t \), which are then transformed into 2D pixel coordinates using the camera view defined by the rotation matrix \(R\) and translation vector \(T\). This transformation effectively maps the 3D points onto the image plane, ensuring alignment with the image's features:
\begin{equation}
p_i = \Pi(R \cdot x_{t,i} + T),
\end{equation}
where \( p_i \) represents the 2D pixel coordinates corresponding to the 3D point \( x_{t,i} \), and \( \Pi(\cdot) \) denotes the rasterization function.

From the rasterized image, the visible points \( V \subseteq \{ p_i \} \) are identified. For each visible pixel \( p_i \in V \), the image feature \( F(I, p_i) \), which represents the feature at pixel \( p_i \) in the image \( I \), is assigned to the corresponding 3D point \( x_{t,i} \). This process ensures that the features from the visible pixels are accurately associated with the 3D points based on their visibility and mapping through the rasterization. This assignment is given by:
\begin{equation}
f(x_{t,i}) = F(I, p_i) \quad \text{for} \quad p_i \in V .
\end{equation}

In this paper, the same rasterization-based approach will be adapted to project image features onto partially denoised 3D points. However, we find that using only image features is inadequate for accurate reconstruction. To enhance edge representation, we incorporate Sobel edge maps to better guide the model.

\subsection{\papername}
When analyzing an aerial image of a single building, background elements such as parked cars or trees can often be observed, contributing depth and additional characteristics to the image. To ensure the model focuses solely on the building, it is essential to incorporate a binary building mask  (created as described in \cref{subsec:dataset_creation}) as an additional condition. This is accomplished by concatenating the building mask values with the image features extracted using the \gls{gl:ViT} \citep{dosovitskiy2020image}, as shown in \cref{fig:architecture}, before projecting them onto the point cloud  using our generated camera pose.

\begin{algorithm}[t]
\footnotesize
\caption{Optimization of Camera Translation Parameters}
\label{alg:camera-params}
\begin{algorithmic}[1]
\State \textbf{Input:} Point Cloud $\mathcal{P}$, Ground Truth Mask $\mathcal{M}_{gt}$, Initial Camera Translation Parameters $\mathbf{T}_0 = (t_{x0}, t_{y0}, t_{z0})$
\State \textbf{Output:} Optimized Camera Translation Parameters $\mathbf{T}^* = (t_x^*, t_y^*, t_z^*)$

\State Initialize camera translation parameters $\mathbf{T} \gets \mathbf{T}_0$

\While{cost improvement \textgreater 1e-4 \text{ and iteration} \textless 1000}

    \State Rasterize point cloud $\mathcal{P}$ using current camera translation parameters $\mathbf{T}$ and fixed rotation parameters $\mathbf{R} = 0$ to generate a rasterized image $\mathcal{M}_r$
    
    \State Compute bounding boxes for the ground truth and rasterized images:
    \State \ \ \ $\text{bbox}_{gt} \gets \text{BoundingBox}(\mathcal{M}_{gt})$
    \State \ \ \ $\text{bbox}_{r} \gets \text{BoundingBox}(\mathcal{M}_r)$
    
    \State \textbf{Step 1:} Optimize \(t_z\) (depth)
    \State Define the cost function for depth optimization:
    \State \ \ \ $\text{cost}_z \gets  \sqrt{\begin{aligned}
        & (x_{\text{min}}^{gt} - x_{\text{min}}^{r})^2 + (x_{\text{max}}^{gt} - x_{\text{max}}^{r})^2 + \\
        & (y_{\text{min}}^{gt} - y_{\text{min}}^{r})^2 + (y_{\text{max}}^{gt} - y_{\text{max}}^{r})^2
    \end{aligned}}$
    \State Optimize \(t_z\) using Powell's method to minimize \(\text{cost}_z\)

    \State \textbf{Step 2:} Optimize \(t_x\) (horizontal translation)
    \State Define the cost function for horizontal translation optimization:
    \State \ \ \ $\text{cost}_x \gets \sqrt{
        (x_{\text{min}}^{gt} - x_{\text{min}}^{r})^2 + (x_{\text{max}}^{gt} - x_{\text{max}}^{r})^2
    }$
    \State Fix \(t_z\) and optimize \(t_x\) using Powell's method to minimize \(\text{cost}_x\)

    \State \textbf{Step 3:} Optimize \(t_y\) (vertical translation)
    \State Define the cost function for vertical translation optimization:
    \State \ \ \ $\text{cost}_y \gets \sqrt{
        (y_{\text{min}}^{gt} - y_{\text{min}}^{r})^2 + (y_{\text{max}}^{gt} - y_{\text{max}}^{r})^2
    }$
    \State Fix \(t_z\) and \(t_x\), and optimize \(t_y\) using Powell's method to minimize \(\text{cost}_y\)
\EndWhile

\State \textbf{Return} optimized camera translation parameters $\mathbf{T}^* = (t_x^*, t_y^*, t_z^*)$
\end{algorithmic}
\end{algorithm}

In addition to the binary building mask, roof edges play a crucial role in reducing ambiguities during the reconstruction process. By providing clear structural boundaries, they help define the building’s silhouette and improve the model’s accuracy. This is especially important for entire building reconstruction, as the accurate representation of roof edges directly influences the overall geometry and completeness of the model. To enhance edge clarity, we also concatenate Sobel-filtered images with the binary mask and the image features, effectively highlighting transitions at the roof's perimeter. The Sobel filter detects sharp intensity changes, capturing critical gradients that outline the roof. This approach enhances the model’s ability to accurately capture the building’s geometry while reducing confusion from surrounding elements and making our model more aware of the building's structure.

In our diffusion network, we implement the \gls{gl:PVCNN} \citep{liu2019point} architecture and employ $\mathcal{L}_2$ loss for supervision, similar to our reference models. By utilizing~\gls{gl:CDPM}, we ensure that the model concentrates on accurate reconstruction of the point cloud rather than attempting to locate the center, enhancing the overall quality of the output.

\subsection{Dataset} \label{subsec:dataset_creation}

\cite{ren2021intuitive} built a dataset using their optimization-based roof reconstruction method, which designs a primal or dual roof graph as input and optimizes the geometry of the roof structure to generate a planar 3D polygonal roof. This dataset consists of complete 3D building meshes paired with aerial images, along with their corresponding vertex and face files.
Using this dataset, we identified several missing components necessary to fit the pipeline. Specifically, we lacked the following: building masks, point cloud ground truth, Sobel edge maps, and camera parameters, which are essential for the rasterization process. In this part, we will explain how we generated each of these components.

\subsubsection*{Building Binary Mask. } 
\begin{figure}[h]
  \centering
  \subcaptionbox{Aerial RGB Image \label{fig:aerialrgb}}
                {\parbox{.23\linewidth}{%
                    \includegraphics[width=\linewidth]{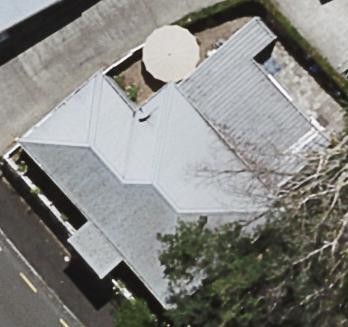}
                   }%
                }
  \subcaptionbox{Binary Mask \label{fig:binarymask}}
                {\parbox{.23\linewidth}{%
                   \includegraphics[width=\linewidth]{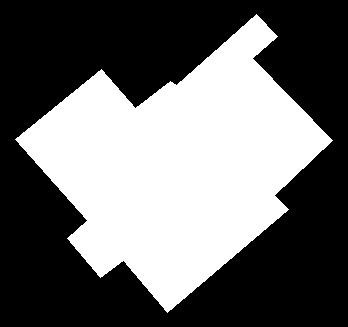}
                 }%
                }
  \subcaptionbox{Sobel Edge Map \label{fig:sobel}}
                {\parbox{.216\linewidth}{%
                   \includegraphics[width=\linewidth]{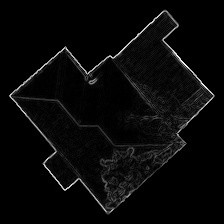}
                 }%
                }
 \subcaptionbox{Point Cloud\label{fig:datapc}}
                {\parbox{.23\linewidth}{%
                   \includegraphics[width=\linewidth]{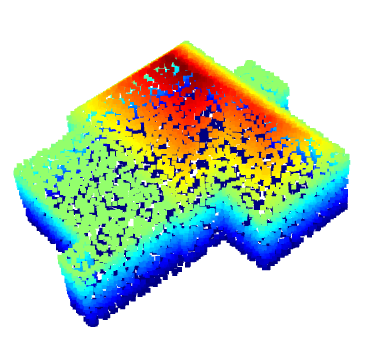}
                 }%
                }                   
  \vspace{0.2cm}
  \caption{%
  	Sample from our dataset, where (\ref{fig:aerialrgb}) represents the RGB aerial image from our initial dataset, (\ref{fig:binarymask}) is the binary mask of the building, and (\ref{fig:sobel}) is the Sobel edge map, while (\ref{fig:datapc}) shows our sampled normalized point cloud.
  }
   \label{fig:datasample}
\end{figure}

The vertex files contain the image coordinates of the roof corner points, while the face files define the sequences of vertex indices that represent the roof segments. Using these coordinates, we create a binary mask for each roof instance, marking the building boundaries and differentiating the structures from the surrounding area. This approach is particularly effective because state-of-the-art image segmentation models, can sometimes obscure parts of the building due to occlusions, such as tree leaves on the roof, resulting in incomplete or inaccurate masks.  For our application, which requires precise camera parameter estimation (since we rely on the binary mask to extract accurate camera parameters) and complete coverage of the building, including occluded areas, it is essential to have accurate masks, as shown in \cref{fig:binarymask}. Additionally, these models can occasionally misinterpret shadows as part of the building, leading to further inaccuracies. Our method ensures that we achieve accurate and comprehensive masks, effectively addressing these potential issues.
\subsubsection*{Point Cloud Generation. } To generate the required point clouds from the 3D meshes, we employed CloudCompare \citep{girardeau2016cloudcompare} to automatically sample points from the triangulated faces of the meshes. This process resulted in a point cloud for each building with a consistent density of 10,000 points, as illustrated in \cref{fig:datapc}.

\subsubsection*{Sobel Edge Map. } To focus exclusively on the building's edges while ignoring those from the background, we first applied a binary mask to isolate the building. After removing the background, the masked image was processed using the Sobel operator, which detects intensity changes in both horizontal and vertical directions to compute the edge map. This results in an image, as shown in \cref{fig:sobel}, where the building's edges are clearly highlighted, capturing the key contours and boundaries.

\subsubsection*{Camera Pose.}
To effectively project 3D point clouds onto 2D images, it is essential to have accurate camera parameters, particularly the camera's position, represented by the translation vector, and the rotation matrix. Since the initial 3D models are consistently oriented relative to the input images, there is no need to optimize the rotation matrix, and we can set it to zero. In this task, our primary focus is on optimizing the translation vector to ensure that the rasterized image of the 3D point cloud aligns with the ground truth mask. When these images align and overlap correctly, the rasterized image will coincide with the actual RGB image, allowing us to accurately project the corresponding features from the RGB image into the point cloud. The challenge lies in determining the correct translation vector that best aligns the rasterized image with the ground truth.

We propose an optimization strategy, detailed in \cref{alg:camera-params}, that adjusts the camera translation parameters \((t_x, t_y, t_z)\) to minimize the misalignment between the rasterized image and the ground truth mask. The cost function is based on the difference between the bounding box of the building in the rasterized image and the corresponding bounding box in the binary mask.

In our optimization process, we first center the foreground region (building) of the binary mask along the \(x\)- and \(y\)-axes. This centering is critical for optimizing the \(z\)-coordinate (depth), as the central alignment ensures that any changes in depth correspond to changes in the bounding box size, rather than the image's position. Without centering, the translation in \(x\)- and \(y\)-axes could falsely appear as depth-related errors. However, for optimizing \(x\) and \(y\), centering is unnecessary since we are directly modifying the horizontal and vertical translations.

We use \textbf{Powell’s method} \citep{10.1093/comjnl/7.2.155} for optimization, as it is well-suited for minimizing non-linear, non-differentiable cost functions like ours. These cost functions compare the bounding boxes of the projected point cloud and the ground truth mask to measure alignment: $\text{cost}_z$ ensures the object’s size matches by computing the Euclidean distance between all four edges (horizontal and vertical), $\text{cost}_x$ aligns the object horizontally by minimizing the distance between the left and right edges, and $\text{cost}_y$ aligns it vertically by minimizing the distance between the top and bottom edges. Powell’s method is particularly effective for this task because it does not require gradient information and instead relies on a series of line searches, making it ideal for our discrete, pixel-based optimization problem.
The sequential optimization strategy involves:
\begin{enumerate}
    \item Optimizing the \(z\)-coordinate (depth).
    \item Fixing \(z\), then optimizing the \(x\)-coordinate (horizontal translation).
    \item Fixing both \(z\) and \(x\), then optimizing the \(y\)-coordinate (vertical translation).
\end{enumerate}
This step-by-step approach ensures accurate alignment between the rasterized image of the 3D point cloud and the 2D ground truth mask.

After obtaining the optimized camera translation parameters, we check the Intersection over Union (IoU) between the bounding boxes of the rasterized image generated with these parameters and the ground truth mask. If the IoU exceeds 93\%, we consider it a valid input for further processing. This ensures that only well-aligned configurations are utilized.

\section{Experiments}\label{sec:Experiments}

\subsubsection*{Implementation Details. }
We trained the diffusion model using a batch size of 8 for a total of 100,000 steps, employing \gls{gl:MAE} for feature extraction. Optimization was handled by the AdamW \citep{adamw} optimizer, which utilized a dynamic learning rate and a warmup schedule. All experiments were carried out on a single GPU to ensure optimal performance.
For dataset preparation, we initially optimized camera parameters. The model was trained on a dataset consisting of 2,544 input images, which were resized to $224 \times 224$ before applying   \cref{alg:camera-params}. Additionally, we employed the Sobel operator to generate edge maps, which provides a simple yet effective method for detecting edges. The Sobel operator uses two $3 \times 3$ convolution masks to estimate gradients in the x- and y-directions, allowing us to highlight building edges that are crucial for accurate 3D reconstruction. 

\begin{figure*}[t]

  \subcaptionbox{Single Input Image \label{fig:input1}}
                {\parbox{.18\linewidth}{%
                    \includegraphics[width=\linewidth]{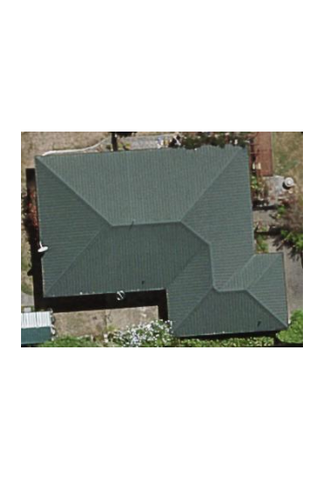}

                    \includegraphics[width=\linewidth]{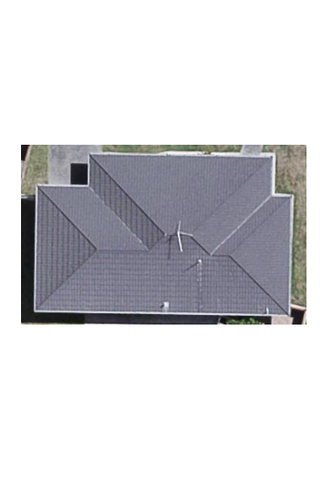}
                   }%
                }
  \subcaptionbox{\pc \label{fig:pc1}}
                {\parbox{.2\linewidth}{%
                   \includegraphics[width=\linewidth]{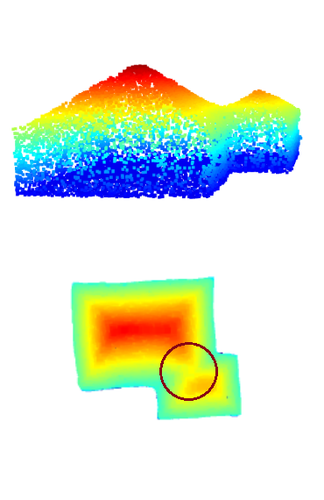}

                   \includegraphics[width=\linewidth]{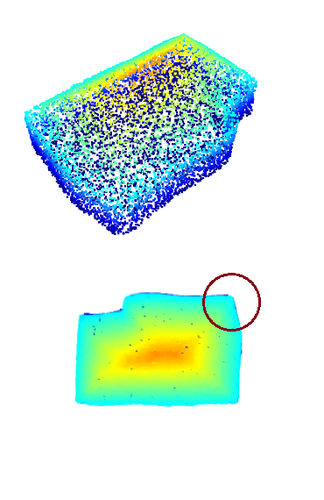}
                 }%
                }
  \subcaptionbox{\ccd \label{fig:ccd3dr1}}
                {\parbox{.2\linewidth}{%
                   \includegraphics[width=\linewidth]{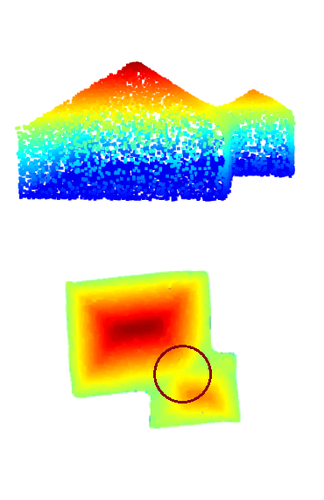}

                     \includegraphics[width=\linewidth]{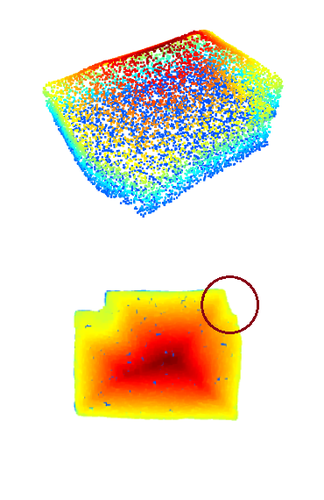}
                 }%
                }
 \subcaptionbox{\papername(ours)\label{fig:ourspred}}
                {\parbox{.2\linewidth}{%
                   \includegraphics[width=\linewidth]{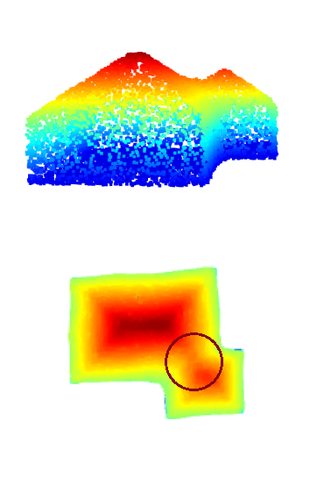}

                   \includegraphics[width=\linewidth]{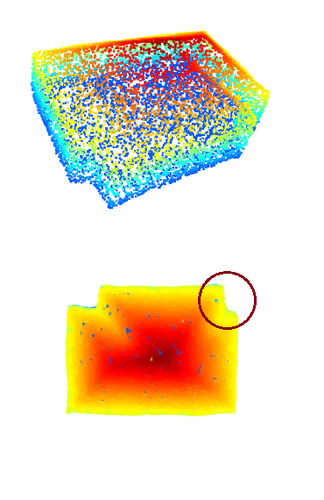}
                 }%
                }    
 \subcaptionbox{Ground Truth\label{fig:gt1}}
                {\parbox{.2\linewidth}{%
                   \includegraphics[width=\linewidth]{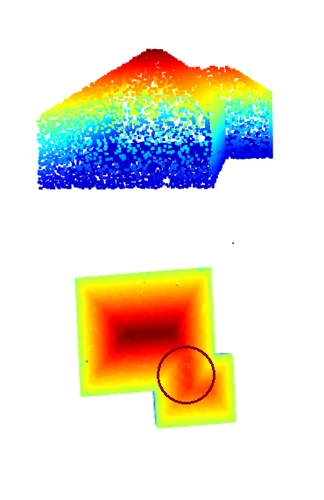}

                   \includegraphics[width=\linewidth]{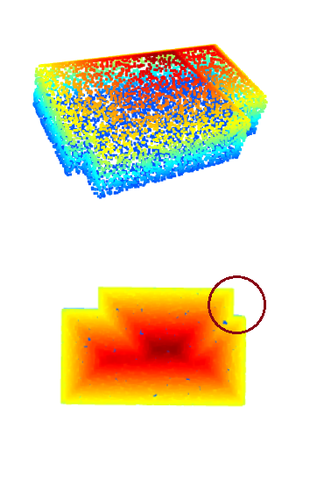}
                 }%

                }\\
                
  \vspace{0.2cm}

  \caption{%
  	Qualitative comparison from a top view and an alternative viewing angle, showcasing our model alongside the baseline models.
 (\ref{fig:input1}) represents the aerial input single-view image, (\ref{fig:pc1}) shows the reconstruction using \pc, (\ref{fig:ccd3dr1}) displays the reconstruction using \ccd, (\ref{fig:ourspred}) illustrates the reconstruction using our method \textsc{\papername}, and (\ref{fig:gt1}) presents the ground truth. The comparison highlights the superior performance of our method in accurately reconstructing the roof's shape. This is evident when looking at the top-view perspective of all models: \papername generates a point cloud with color distribution that closely matches the ground truth. In contrast, \pc not only underestimates the roof height but also misses certain sections, while \ccd shows incomplete roof reconstructions with missing key details, as highlighted in red circles. These edge and structural issues become even more apparent when verified from alternative viewing angles.
  }
   \label{fig:pred1}
\end{figure*}
For point cloud data, we used point clouds containing 10,000  points, as a higher number of points improves the level of detail in the reconstructed structures.
To ensure consistency across all buildings in our dataset, we normalized the point cloud data to a unit sphere. This normalization is crucial for applying  (\cref{alg:camera-params}) consistently across all buildings, as it ensures that all buildings share the same scale. Without this step, the initial camera translation parameters \( t_{x0} \), \( t_{y0} \), and \( t_{z0} \) would need to be manually adjusted for each building, leading to an inconsistent and inefficient process. However, while this normalization simplifies the process, it does result in the loss of actual height values for the buildings.
\begin{figure}[t]
\centering
  \subcaptionbox{Single Input Image \label{fig:singlequalitativeresult}}
                {\parbox{.32\linewidth }{%
                    \includegraphics[width=\linewidth]{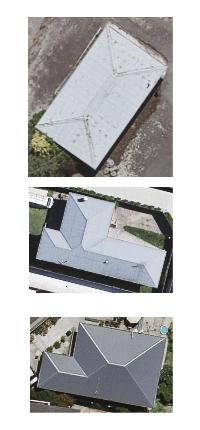}
                }%
                }
  \subcaptionbox{\papername View 1  \label{fig:novelview}}
                {\parbox{.32\linewidth}{%
                   \includegraphics[width=\linewidth]{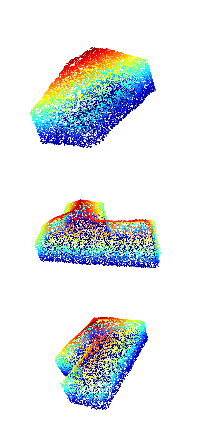}
                 }%
                }
  \subcaptionbox{\papername View 2 \label{fig:novelview2}}
                {\parbox{.32\linewidth}{%
                   \includegraphics[width=\linewidth]{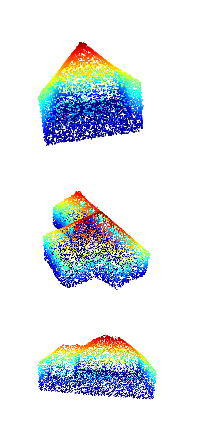}
                 }%
                }                
                \\
  \vspace{0.2cm}

  \caption{%
    Qualitative results of two novel reconstructed views (\ref{fig:novelview}) and (\ref{fig:novelview2}) generated by \papername from single input aerial images (\ref{fig:singlequalitativeresult}). 
    This demonstrates the capability of our solution to reconstruct the complete point cloud of the entire building. }
   \label{fig:qual1}
\end{figure}
The optimized camera parameters (\( t_x^* \), \( t_y^* \), and \( t_z^* \)) generally fall within the range of 0 to 1 due to the unit sphere normalization.  For simplicity and broader applicability, during the inference stage, we set the default camera parameters \( t_x = 0 \), \( t_y = 0 \), and \( t_z = 1 \), simulating a top-down view along the Z-axis. This approach allows for generating reconstructions without needing to specify precise camera parameters for each individual building.

\subsubsection*{Quantitative Results. }We evaluate point cloud reconstructions using \gls{gl:CD} and F-Score@0.001. \Gls{gl:CD} quantifies the similarity between two point clouds by measuring the spatial proximity of their points. It identifies the nearest ground truth point for each point in the predicted cloud and vice versa, averaging these minimum distances to provide an overall measure of alignment. However, \gls{gl:CD}'s sensitivity to outliers can skew the average distance, leading to potentially misleading assessments of reconstruction quality. Therefore, while \gls{gl:CD} is a valuable metric for evaluating point cloud accuracy, it is beneficial to use additional metrics for a more comprehensive analysis.

In this context, we also utilize the F-Score, which defines a point as correctly predicted only if it lies within a specified threshold of its nearest match. For point clouds containing 10,000 normalized points (within the range of 0 to 1), we set this threshold to 0.001. This small threshold ensures that a point is considered correctly predicted only if it is within 0.1\% of its nearest reference point, allowing us to detect even minor discrepancies and facilitating a precise assessment of reconstruction quality.

In \cref{tab:results}, we compare our method to baseline approaches by focusing on the best average performance of the metrics. Notably, the baseline models do not utilize Sobel edge maps as conditions, while \papername integrates Sobel edge maps as an additional condition, enhancing the overall performance. Specifically, we observe a 21.87\% increase in F-Score compared to the \pc method, and a 4.83\% increase over the \ccd model. In terms of \gls{gl:CD}, our approach achieves a 10.44\% better performance compared to \pc and 3.40\% better performance relative to \ccd. These results demonstrate the efficiency of our method in producing more precise and dependable point cloud reconstructions.
\begin{table}[t!]
\centering

\resizebox{8cm}{!}{
\begin{tabular}{lccc}
\toprule
\textbf{Method} & \textbf{F-Score @0.001} $\uparrow$ & \textbf{CD ($\times 10^{-3}$)} $\downarrow$ \\
\midrule
\pc & 0.535 & 3.172 \\
\midrule
\ccd & 0.622 & 2.941 \\
\midrule
\papername  (Ours) & \textbf{0.652} & \textbf{2.841} \\
\bottomrule
\end{tabular}
}
\caption{Quantitative results for different methods are presented under Chamfer Distance (CD, reported as $\times 10^{-3}$) and F-Score@0.001, demonstrating that our method outperforms the others.}
\label{tab:results}
\end{table}

\subsubsection*{Qualitative Results. }

\cref{fig:qual1} illustrates the qualitative results of our proposed approach, highlighting its effectiveness in reconstructing 3D point clouds that accurately capture structural details and can generate fully complete buildings from multiple viewing angles, rather than just a single perspective.
In \cref{fig:pred1}, we present comparative visualizations of the \pc and \ccd models alongside \papername. Each model’s point cloud is displayed from both a top-down perspective and an additional viewing angle to facilitate a comprehensive analysis. The color mapping of the point clouds, which represents height variations along the Z-axis, is critical for visualization. This color representation plays a key role in comparing the predicted heights against ground truth data, allowing for a clear assessment of the performance of various prediction methods.
When visually comparing our method to others, it becomes evident that our predictions capture finer details, particularly around roof edges and their impact on the overall building structure as shown in red circles in the figure. Additionally, our approach provides more accurate elevation values, offering a distinct advantage in representing complex surface features. While the point clouds have been normalized, visualizing them is crucial for distinguishing these differences in elevation representation across the models.

\glsresetall
\section{Conclusion}
In this paper, we achieved the reconstruction of complete 3D building point cloud from single-view aerial images, generating point clouds that encompass not only rooftops but also walls and ground surfaces. By utilizing edge maps as additional conditions, our approach yields a more detailed and precise reconstruction. Furthermore, we developed a new dataset containing fully normalized point clouds along with camera poses that enable the point clouds to accurately overlap with the single RGB images during projection. The effectiveness of our approach is affirmed through both qualitative and quantitative results, which demonstrate significant improvements in capturing critical structural details compared to existing methods.  This research paves the way for cost-effective and scalable 3D building reconstruction using single aerial images. In future work, we will incorporate actual height values to enhance the utility of our models. Additionally, we will expand our focus to entire building scene reconstruction, as the success achieved in reconstructing single buildings lays a promising groundwork for addressing the complexities involved in modeling complete urban environments.
\bibliography{bibliography}

\end{document}